# SUPERVISED ADAPTIVE THRESHOLD NETWORK FOR INSTANCE SEGMENTATION

*Kuikun Liu, Jie Yang, Cai Sun, Haoyuan Chi*

Shanghai Jiao Tong University, China

## ABSTRACT

Currently, instance segmentation is attracting more and more attention in machine learning region. However, there exists some defects on the information propagation in previous Mask R-CNN and other network models. In this paper, we propose supervised adaptive threshold network for instance segmentation. Specifically, we adopt the Mask R-CNN method based on adaptive threshold, and by establishing a layered adaptive network structure, it performs adaptive binarization on the probability graph generated by Mask R-CNN to obtain better segmentation effect and reduce the error rate. At the same time, an adaptive feature pool is designed to make the transmission between different layers of the network more accurate and effective, reduce the loss in the process of feature transmission, and further improve the mask method. Experiments on benchmark data sets indicate that the effectiveness of the proposed model.

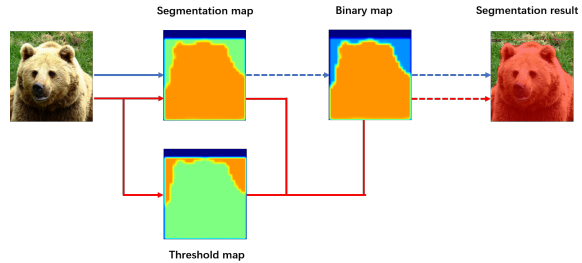

**Fig. 1**. Traditional pipeline (blue flow) and our pipeline (red flow). Dashed arrows are the inference only operators; solid arrows indicate differentiable operators in both training and inference.

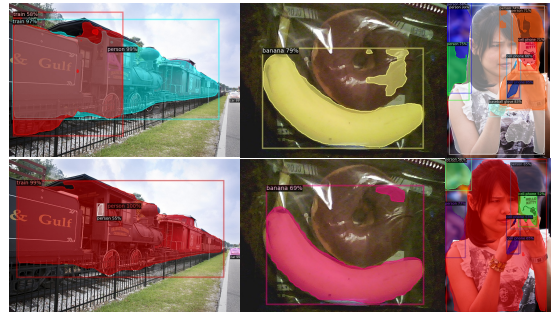

**Fig. 2**. First row: Selected cases of coarse boundaries appeared in the instance segmentation results of Mask R-CNN. Second row: Our proposed method can predict more precise boundaries and less misclassification rate.

## 1. INTRODUCTION

Instance segmentation is a very basic but important task in computer vision area. Its goal is to predict class label and pixel instance masks that render different Numbers of instances in localized images. The task has broad applications in areas such as self-driving cars, video surveillance and intelligent manufacturing. It needs to distinguish the location of each instance and the pixel contained in each instance at the same time, which can be regarded as a combination of object detection and semantic segmentation. With the rapid development of convolutional neural networks, there have been many state-of-art networks for instance segmentation. A large number of these networks are based on object detection networks. The location information of the instance is provided by the object detection network, and the semantic segmentation branch is added to determine all pixels contained in the instance. Among them, Mask R-CNN [1] is the most classic and successful one. Mask R-CNN combined Faster R-CNN [2] and FCN to realize the baseline of instance segmentation.

Mask R-CNN uses the method shown in Fig 1 (blue dotted line) to obtain the final segmentation results: They set a fixed threshold for converting the probability graph generated by FCN into binary images. Mask R-CNN treats all pixels in the probability graph equally, ignoring the different probability distributions generated by different positions, like boundary, background and foreground. There are also misclassified and difficultly classified pixels, such as boundary. Our approach (following the red arrows in Fig 1) is designed to insert binarization operations into segmented networks for joint optimization. In this way, the threshold value of each position of the image can be predicted adaptively, so that the pixel can be completely distinguished from the foreground and background, reducing the probability of misclassification and obtaining a clearer segmentation image.

The comparison results are shown in Fig 1. Compared with the addition of adaptive threshold (the second row), traditional Mask R-CNN (the first row) tends to output rough and fuzzy segmentation results, and there is an illogical overlap between objects, as well as a large number of misclassified cases. The main contributions of this paper are as follows: (1) we firstly explore the possibility of finding out the relationship between different thresholds and segmentation results, and design an end-to-end way than can optimize threshold map in the field of instance segmentation. (2) We propose a novel adaptive threshold mask R-CNN (ATMask R-CNN), which is the first work to improve the mask level positioning accuracy by explicitly using the threshold information in the Mask R-CNN framework. (3) The ATMask R-CNN is conceptually simple but effective. Without bells and whittles, ATMask R-CNN performed 1.7AP better than Mask R-CNN on COCO val set.

## 2. RELATED WORK

The existing instance segmentation methods can be divided into two types: detection-based approach and segmentation-based approach. The detection-based approach is generally to use the detector to obtain candidate regions, then pool or align ROI, and finally obtain mask prediction results. Mask-RCNN [1] adds a new segmentation branch on the basis of the original classification branch and positioning branch, and expands the Faster R-CNN [2], showing a competitive advantage in object detection and instance segmentation. Based on Mask R-CNN, PANet [5] added bottom-up path enhancement. In order to shorten the information transmission path and utilize the precise positioning information of low-level features, it proposed dynamic feature pooling. Each proposal uses the features of all layers of the pyramid to avoid arbitrary allocation of proposals. MaskLab [6] is built on Faster R-CNN. After detecting the target box, the corresponding semantic channel is selected and clipped using the corresponding category, and then the segmentation mask is obtained through convolution of 1x1 combined with direction prediction. In [7], based on Mask R-CNN, a network block is added to predict the quality of the sample mask, so as to solve the mismatch between mask score and mask quality. HTC [8] has designed a multi-tasking multi-stage hybrid cascade structure and integrated a branch of semantic segmentation to enhance spatial context to achieve higher precision segmentation results.

The segmentation-based approach first uses the pixel-level segmentation on the image and then groups the pixels for each object. BIS [9] uses CRF [10] to find instances. SIS [11] introduces metric learning and completes subdivision through clustering. SOIS [12] uses semi-convolution to split instances. InstanceCut [13] outputs the semantic split diagram and boundaries for all instances and then uses Multi-Cut to split the final instance. SGN [14] proposed to solve the task of instance segmentation by a series of neural networks, in which each neural network is to aggregate semantic information at this level, so as to gradually construct object instances using simple structures.

## 3. PROPOSED MODEL

In this part, we will introduce the network structure of our proposed adaptive threshold Mask R-CNN, how to add the adaptive threshold branch to Mask R-CNN, as well as its training method and loss function. The overall structure of the model is shown in Fig 3. Next, we will discuss it in detail.

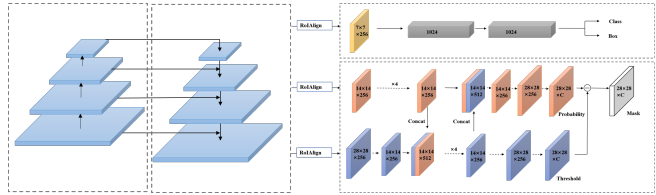

**Fig. 3**. The Overall architecture of Adaptive-Threshold Mask R-CNN (ATMask R-CNN). The dotted arrow denotes 3 × 3 convolution and the solid arrow denotes identity connection unless specified annotation in adaptive-threshold mask head. "×4" denotes a stack of four consecutive convs.

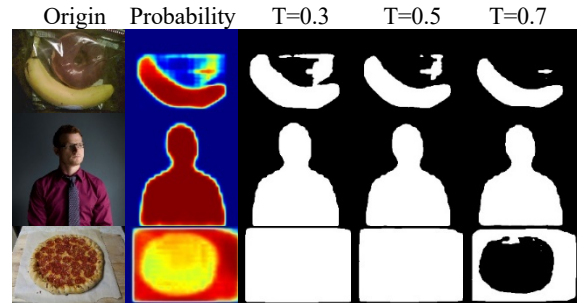

**Fig. 4**. Visualization of some predicted masks probability heat map of Mask R-CNN and mask results with different threshold (0.3, 0.5, 0.7).

### 3.1. Motivation

In Mask R-CNN, the essence of segmentation branch is to dichotomy each pixel. Different position has different distributions. There are also misclassified and difficultly classified pixels, such as boundary. For better understanding the problem, we visualize the probability maps generated by segmentation branch with Mask R-CNN, and show the final segmentation results generated by the probability graphs under different threshold values. In Fig. 4, we can observe that there are a large number of misclassified pixels, and the probability value of boundary points is erratic. At the same time, by setting different thresholds, the final segmentation result and misclassification will be greatly changed. This means that threshold can not only affect the precision of segmentation, but also change the misclassified situation. Thus, it can be seen that setting different thresholds for different positions can indeed improve the segmentation effect. Therefore, we proposed an adaptive threshold branch.

### 3.2. Binarization

For the probability graph P with height H and width W generated by the segmentation branch, it will finally be converted into a binarization graph. In the binarization graph, 1 represents the foreground and 0 represents the background, which can be represented by the following formula.

$$B_{i,j} = f(x) = \begin{cases} 1, & if\ P_{i,j} \geq t \\ 0, & otherwise \end{cases} \quad (1)$$

where $t$ represents the threshold value of the preset value, and $P_{i,j}$ represents the probability value of the probability graph at the point (i, j). Adaptive threshold branch also produces a threshold graph with height H and width W. The standard binarization described in equation 1 does not allow gradient descent. Therefore, it cannot be optimized with the segmentation network during training. We use the proposed differentiable binarization rcnn:

$$B_{i,j} = \frac{1}{1+e^{-k(P_{i,j}-T_{i,j})}} \quad (2)$$

where $T_{i,j}$ represents the threshold value of the threshold graph at (i, j), and $B_{i,j}$ represents the binarization result at (i, j). k is set to 2 according to the experiment.

### 3.3. Adaptive Threshold

The adaptive threshold branch and the segmentation branch learn the probability graph and threshold graph of the object together in an end-to-end manner. It should be noted that the probability graph is closely related to the threshold graph. The threshold graph needs to use the probability graph to set the threshold from the adaptive threshold graph. Meanwhile, the adaptive threshold graph obtained can feed back the foreground background information of the probability graph to promote the generation of better probability graph. Finally, the obtained probability graph and threshold graph are combined with differentiable binarization module to obtain target segmentation results.

**RoI Feature Extraction:** Rp and Rt are respectively defined as RoI features and used for probabilistic prediction and threshold prediction. After [15], Rp is generally extracted from a specific feature pyramid (P2 ~ P5) according to the size of the proposed area. For Rt we obtained from P2, the highest resolution feature level, which contains more underlying information. FPN Then, by sampling it down with 3 by 3 convolution, the output feature is denoted as Rt. Rt has the same resolution as Rp and is used for feature fusion. The feature fusion scheme in ATMask R-CNN is shown in Fig 3. The probability RoI feature Rp is fed into four consecutive 3×3 convolutions and the feature layer generated by Rt is fused, and then fed back to the feature generated by Rp after four convolutions.

**Probability→Threshold (P2T) Fusion:** The adaptive threshold network needs to obtain probability results so as to generate the adaptive threshold. Since the target area value of the probability graph is relatively high and the background area value is relatively low, while the target area value of the threshold graph is relatively low and the background area value is relatively high, reasonable results cannot be obtained if summation is used to achieve fusion, so we select the feature merging layer.

$$F_B = \text{Concat}(F_P, F_T) \quad (3)$$

where $F_P$ represents threshold feature, $F_T$ represents probability feature, and $F_B$ denotes the final segmentation feature map.

**Threshold→ Probability (T2P) Fusion:** We fuse the threshold feature with the probability feature for enriching the probability features by providing foreground and background information. The fusion block is the same as the P2T fusion block.

### 3.4. Learn and Optimization

For probability graphs, threshold graphs and final binary graphs generated by threshold graphs and probability graphs, binary cross-entropy (BCE) loss is used.

$$L_{probability} = L_{threshold} = L_{binary}$$
$$= -\sum_i [(1-y_i)\log(1-p_i) + y_i \log p_i] \quad (4)$$

where $L_{threshold}$ represents threshold loss, $L_{probability}$ represents probability loss, and $L_{binary}$ denotes the final segmentation feature loss.

Multitasking learning has been shown to be effective in many works, and it performs better on different tasks than training alone. Since the probability graph and threshold graph are cross-linked by two fusion blocks, joint training can enhance the feature representation of probability and threshold prediction. We define a multi-task loss, as shown below:

$$L = L_{cls} + L_{box} + L_{probability} + L_{threshold} + \lambda L_{binary} \quad (5)$$

The classification loss $L_{cls}$ and regression loss $L_{box}$ were inherited from Mask R-CNN. λ is set to 2.

## 4. EXPERIMENT AND DISCUSSION

We conducted extensive experiments on the challenging COCO data set [16] to demonstrate the effectiveness of Adaptive-Threshold Mask R-CNN. In order to better understand each component of our approach, we provided detailed ablation experiments on COCO.

**Experimental Details:** We used Mask R-CNN as the benchmark and developed on this basis. All the hyperparameters remain the same. Unless otherwise specified, we shall treat RESnet-50 and FPN as backbones. We used ImageNet's pre-trained weights to initialize our backbone network. In accordance with standard practice, we used Synchronized SGD to train all models on four NVIDIA GPU2080TI with initial learning rates of 0.02 and 12 images per small batch, respectively, for 90,000 iterations, and reduced learning rates by 0.1 and 0.01 times iterations, and reduced learning rates by 0.1 and 0.01 times after 60,000 and 80,000 iterations. For the backbone of RESnet-101 and FPN, due to GPU memory, we initialize our network with pre-training weights based on Faster R-CNN, freeze classification and detection branches, and the initial learning rate is 0.02 and 8 images per small batch respectively, carry out 50,000 iterations, and reduce the learning rate by 0.1 and 0.01 times after 30,000 and 40,000 iterations respectively.

Table 1: Comparison with state-of-the-art methods for instance segmentation on COCO test-dev2017(* denotes our implementation)

| Method | Backbone | AP AP50 AP75 | APS APM APL |
|---|---|---|---|
| Mask R-CNN | ResNet-101-FPN | 35.7 58.0 37.8 | 15.5 38.1 52.4 |
| Mask R-CNN | ResNeXt-101-FPN | 37.1 60.0 39.4 | 16.9 39.9 53.5 |
| MaskLab | ResNet-101-FPN | 35.4 57.4 37.4 | 16.9 38.3 49.2 |
| MaskLab+ | ResNet-101-FPN | 37.3 59.8 39.6 | 19.1 40.5 50.6 |
| Mask Scoring R-CNN | ResNet-50-FPN | 35.8 56.5 38.4 | 16.2 37.4 51.0 |
| Mask Scoring R-CNN | ResNet-101-FPN | 37.5 58.7 40.2 | 17.2 39.5 53.0 |
| CondInst [17] | ResNet-50-FPN | 35.4 56.4 37.6 | 18.4 37.9 46.9 |
| BlendMask [3] | ResNet-50-FPN | 34.3 55.4 36.6 | 14.9 36.4 48.9 |
| PointRend [4] | ResNet-50-FPN | 36.3  -   - | -  -  - |
| Mask R-CNN* | ResNet-50-FPN | 35.0 56.1 37.3 | 18.8 37.0 46.1 |
| ATMask R-CNN | ResNet-50-FPN | 36.1 56.6 39.1 | 18.4 38.2 47.1 |
| Mask R-CNN* | ResNet-101-FPN | 35.9 59.2 37.8 | 20.4 38.4 46.4 |
| ATMask R-CNN | ResNet-101-FPN | 37.8 60.1 40.5 | 21.2 40.3 49.3 |

## 4.1. Overall Results

We first evaluated the ATMask R-CNN with different backbone on COCO and compared it with the Mask R-CNN. As shown in Table 2, our method is superior to Mask R-CNN with the same backbone input. Compared with Mask R-CNN, ATMask R-CNN with ResNet-50-FPN and ResNet-101-FPN get 1.3AP and 1.6AP improvement, respectively. Therefore, the adaptive threshold branch can effectively improve the segmentation results of the prediction.

In Table 1, we compare ATMask R-CNN with some of the latest instance segmentation methods. All models were trained on COCO Train2017 and evaluated on COCO Test-dev2017. It gives us an unforgettable impression.

## 4.2. Ablation Experiment

In order to understand the working principle of ATMask R-CNN, we conducted detailed experiments to analyze the components in ATMask R-CNN. Table 3 shows the results of gradually adding components to the Mask R-CNN baseline. Each component of our proposed ATMask R-CNN will be studied in the following sections.

**RoI Feature Extraction:** Based on the probability graph, we have already contained a large amount of high-level information, and the probability at the boundary is more erratic. Therefore, the lower level information is needed to assist the setting of the threshold. Meanwhile, the introduction of the lower level information can also improve the precision of segmentation. Therefore, we adopted the following strategy: change the source of the ROI feature of Rt. Table 4 shows that it is more efficient to derive ROI features from P2 from the threshold branch.

**Feature Fusion:** In Section 3.3, we show the relationship between the threshold graph and the probability graph. The fusion between threshold branch and probability branch will enrich the representation of each characteristic layer. Table 5 shows more results: If there is no fusion, Mask R-CNN can be improved by 0.2AP, which is the benefit of multi-task learning. With P2T and T2P fusion modules, ATMask R-CNN improves 1.1AP over Mask R-CNN.

Table 2. Comparison with Mask R-CNN on COCO val2017

| Method | Backbone | AP | $AP_{50}$ | $AP_{75}$ |
|---|---|---|---|---|
| Mask R-CNN | ResNet-50-FPN | 34.8 | 55.6 | 37.3 |
| ATMask R-CNN | ResNet-50-FPN | 36.1 | 56.2 | 39.0 |
| Mask R-CNN | ResNet-101-FPN | 35.9 | 58.6 | 38.2 |
| ATMask R-CNN | ResNet-101-FPN | 37.5 | 59.4 | 40.4 |

Table 3. Experiment results on COCO val2017 of adding components to Mask R-CNN. We gradually add threshold branch, fusions between probability and threshold features, and our RoI feature extraction strategy for threshold features. (k=1)

| Threshold | Fusion | RoI Strategy | AP | $AP_{50}$ | $AP_{75}$ | $AP^b$ |
|---|---|---|---|---|---|---|
| - | - | - | 34.8 | 55.6 | 37.3 | 38.1 |
| ✓ |  |  | 35.0 | 55.6 | 37.9 | 38.1 |
| ✓ | ✓ |  | 35.7 | 56.2 | 38.5 | 38.1 |
| ✓ | ✓ | ✓ | 35.9 | 56.2 | 38.9 | 38.1 |

Table 4. Experiment results on COCO val2017 for different RoI feature extraction strategies. (k=1)

| Source | AP | $AP_{50}$ | $AP_{75}$ | $AP^b$ |
|---|---|---|---|---|
| P2 | 35.9 | 56.2 | 38.9 | 38.1 |
| P2~P5 | 35.7 | 56.2 | 38.5 | 38.1 |

Table 5. Experiment results on COCO val2017 for the impacts of fusion blocks, i.e. P2T fusion and T2P fusion. (k=1)

| P2T Fusion | T2P Fusion | AP | $AP_{50}$ | $AP_{75}$ | $AP^b$ |
|---|---|---|---|---|---|
| ✗ | ✗ | 35.0 | 55.6 | 37.9 | 38.1 |
| ✓ | ✗ | 35.5 | 56.2 | 38.3 | 38.2 |
| ✓ | ✓ | 35.7 | 56.2 | 38.5 | 38.1 |

## 5. CONCLUSION

In this paper, we improve Mask R-CNN by adding adaptive threshold branch. It reduces the probability of misclassifycation and obtains better segmentation effect. Without bells and whistles, ATMask R-CNN achieves significant and stable improvement on COCO. We hope it will serve as a strong baseline and provide inspiration for this basic research topic.

# 6. REFERENCES

[1] He, K., Gkioxari, G., Doll´ar, P., Girshick, R.B., "Mask R-CNN," *ICCV*, 2017.

[2] Ren, S., He, K., Girshick, R.B., Sun, J., "Faster R-CNN: towards real-time object detection with region proposal networks," *IEEE Trans*. Pattern Anal. Mach. Intell. 39(6), 1137–1149, 2017.

[3] Chen, H., Sun, K., Tian, Z., Shen, C., Huang, Y., Yan, Y., "Blendmask: Top-down meets bottom-up for instance segmentation," *CVPR*, 2020.

[3] Kirillov, A., Wu, Y., He, K., Girshick, R., "Pointrend: Image segmentation as rendering," *CVPR*, 2020.

[5] Liu, S., Qi, L., Qin, H., Shi, J., Jia, J., "Path aggregation network for instance segmentation," *CVPR*, pp. 8759–8768, 2018.

[6] Chen, L., Hermans, A., Papandreou, G., Schroff, F., Wang, P., Adam, H., "Masklab: Instance segmentation by refining object detection with semantic and direction features," *CVPR*, pp. 4013–4022, 2018.

[7] Huang, Z., Huang, L., Gong, Y., Huang, C., Wang, X., "Mask scoring R-CNN," *CVPR*, pp. 6409–6418, 2019.

[8] Chen, K., Pang, J., Wang, J., Xiong, Y., Li, X., Sun, S., Feng, W., Liu, Z., Shi, J., Ouyang, W., Loy, C.C., Lin, D., "Hybrid task cascade for instance segmentation," *CVPR*, pp. 4974–4983, 2019.

[9] Arnab, A., Torr, P.H.S., "Bottom-up instance segmentation using deep higher-order crfs," *BMVC*. BMVA Press, 2016.

[10] Lafferty, J.D., McCallum, A., Pereira, F.C.N., "Conditional random fields: Probabilistic models for segmenting and labeling sequence data," *ICML*, pp. 282–289, 2001.

[11] Brabandere, B.D., Neven, D., Gool, L.V., "Semantic instance segmentation with a discriminative loss function," *CoRR* abs/1708.02551, 2017.

[12] Novotn´y, D., Albanie, S., Larlus, D., Vedaldi, A., "Semi-convolutional operators for instance segmentation," *ECCV*, Lecture Notes in Computer Science, vol. 11205, pp. 89–105, Springer, 2018.

[13] Kirillov, A., Levinkov, E., Andres, B., Savchynskyy, B., Rother, C., "Instancecut: From edges to instances with multicut," *CVPR*, pp. 7322–7331, 2017.

[14] Liu, S., Jia, J., Fidler, S., Urtasun, R., "SGN: sequential grouping networks for instance segmentation," *ICCV*, pp. 3516–3524, 2017.

[15] Lin, T., Doll´ar, P., Girshick, R.B., He, K., Hariharan, B., Belongie, S.J., "Feature pyramid networks for object detection," *CVPR*, 2017.

[16] Lin, T., Maire, M., Belongie, S.J., Hays, J., Perona, P., Ramanan, D., Doll´ar, P., Zitnick, C.L., "Microsoft COCO: common objects in context," *ECCV*, 2014.

[17] Tian, Z., Shen, C., Chen, H., "Conditional convolutions for instance segmentation," *ECCV*, 2020.